\documentclass{article}
\usepackage[T1]{fontenc}
\usepackage[preprint]{corl_2026}
\usepackage{graphicx}
\usepackage{booktabs}
\usepackage{longtable}
\usepackage{array}
\usepackage{amsmath}

\newcommand{\model}{GR00T~N1.7}
\usepackage{xurl}

\hypersetup{pdftitle={Meaning Reaches Object Choice but Not Completion: Where Is a VLA Vulnerable to Landmark Synonyms?}, pdfauthor={Ryoichi Kurihara, Kota Sunamoto}, pdfsubject={}}
\title{Meaning Reaches Object Choice but Not Completion:\\ Where Is a VLA Vulnerable\\ to Landmark Synonyms?}

\author{Ryoichi Kurihara \quad Kota Sunamoto\\ MAMEZO Co., Ltd., Tokyo, Japan}

\begin{document}
\maketitle

\begin{abstract}
VLA policies are fragile to rewording. Success rates alone cannot say whether a policy fails to
choose the object or to complete the task, nor where in the policy to intervene. We study the released
GR00T N1.7 on two LIBERO-Spatial tasks differing only in the landmark word telling which of two bowls
to take. Synonyms absent from the training instructions send the policy to the bowl they name, unlike
made-up words, yet complete 67.6\% of one task's closed-loop episodes, against the original word's
100\%. Before the connector between the vision-language model and the action generator, replacing only
the landmark tokens' vectors with the other landmark's (pasting) sends nearly every episode to the other
bowl. After the connector, pasting at the landmark, the remaining instruction, or the image tokens alone rarely
moves the choice, but any two do. Hiding the landmark tokens from the action generator after the
connector still lowers completion for the synonyms that fall short. Finetuning on diversified landmark
words repairs that one task's completion for synonyms absent from finetuning; finetuning on the original
instructions does not. Training only the connector on the diversified words suffices, as does copying
the connector of a policy fully finetuned on them into one finetuned on the original instructions. An open-loop score on
starting frames misses that weights outside the connector also carry part of the repair. For now, on this evidence, robustness to rewording should be judged in closed loop, with actions executed, intervening on both sides of the connector, scoring choice and completion separately.
\end{abstract}


\section{Introduction}\label{sec:intro}
Vision-language-action (VLA) policies fail when an instruction is reworded without changing the
task \citep{liberopara}, the released \model{} among them \citep{industrialvla}. Yet for some policies,
OpenVLA-OFT above all, success barely changes when the instruction is rewritten or removed, which
LIBERO-Plus reads as the instruction being ignored \citep{liberoplus}. Success rates cannot tell a wrong
choice of object from a failure to finish after the right one. Prior work finds an instruction's
meaning kept inside the policy \citep{gsr,encodednotcontrol} and the intended object kept while success
falls \citep{intact}; what it leaves open
is where inside the policy a single word's effect can be changed (\S\ref{sec:related}).

On one released policy and two tasks that differ only in the landmark word telling which of two
bowls to pick up, we ask where along the path from language to action a single word's \emph{action gap}
can be changed. By that gap we mean the difference between how the policy acts on a wording and how the \emph{parent VLM} resolves it (\S\ref{sec:reach}). The parent VLM is the pretrained vision-language model (VLM) the policy's own VLM was cut from. Our answer is the connector between the VLM and
the action-generating diffusion transformer: the landmark word's tokens act differently before and
after it, and its weights alone can carry a repair, which other weights also partly carry
(claims~2--3). This locates where interventions and repairs act; where the failure originates we leave
open (\S\ref{sec:entry}, \S\ref{sec:position}).

Operationally, we study the \emph{mirror pair} of LIBERO-Spatial \citep{liu2023libero}, two tasks in one scene (Appendix~\ref{app:cells}, Fig.~\ref{fig:scene}): t7
(the bowl on the stove) and t9 (the bowl on the wooden cabinet). We vary the landmark noun, not the verb, which the training instructions share and whose
rewriting costs little (\S\ref{sec:calib}); object nouns also drive most of LIBERO-Para's drop \citep{liberopara}. By a \emph{synonym} we mean a word we judged to name the same landmark in this scene (five per landmark, Table~\ref{tab:subst_cells}). An \emph{unpresented} synonym is one absent from the training instructions,
whether or not pretraining saw it. We say the meaning of a landmark word \emph{reaches
the choice} of bowl---the bowl the policy comes within 10~cm of (\S\ref{sec:stats})---if, for an unpresented synonym, (a) the policy approaches the bowl the synonym
names, (b) a nonce (made-up) word in the same slot rarely approaches that bowl, and (c) an unpresented synonym of the other landmark sends the policy to the
other bowl. We say it does \emph{not reach completion} if (d) its completion is separated ($|z|\ge3$; \S\ref{sec:stats})
below the training word's; otherwise the comparison is unresolved.

The policy's VLM passes one vector per token (the image tokens of two views, the instruction's other
tokens and one to four tokens of the landmark word; \S\ref{sec:path}) to the \emph{action head}, the
model's term for everything after the VLM: a \emph{connector} of 66 tensors that mixes all positions,
then a diffusion transformer that outputs actions. A \emph{position} is an index in this sequence, not a
place in the scene; the \emph{landmark position} is the landmark word's token span. To \emph{paste} is
to replace the vectors at chosen positions with those computed for the same frame under another
instruction (\S\ref{sec:interventions}); to \emph{mask} is to drop positions from the instruction cross-attention
of the action head; a \emph{transplant} copies tensors from one finetuned policy into another without training.

We contribute four claims about this one checkpoint and task pair, each resting on single-change comparisons on the same frozen VLM.
(1)~\textbf{Meaning reaches the choice, not completion}: (a)--(c) hold, and (d) holds on t9's five synonyms pooled and on one
word of t7; elsewhere on t7 completion is unresolved (\S\ref{sec:reach}).

(2)~\textbf{Before the connector the landmark position alone suffices to set the choice; after it the
choice is distributed, yet the landmark position still matters} (\S\ref{sec:preneck}--\S\ref{sec:postneck}, \S\ref{sec:residual}):
the other landmark's vectors pasted there move nearly every episode to the other bowl before the connector and
almost none after it, where any two of the three groups (landmark, rest of the instruction, image) move most episodes. Masking that position after the
connector still costs approach and completion for the synonyms that fall short (\S\ref{sec:stats}).

(3)~\textbf{Finetuning on a few landmark-diversified instructions repairs t9's completion for synonyms
absent from that finetuning, and the connector's 66 tensors suffice for the repair} (\S\ref{sec:var}--\S\ref{sec:transplant}): in three training seeds,
training only those tensors on those instructions suffices, as does copying them from the landmark-diversified policy into one finetuned on the original instructions; copying the other
action-head tensors instead carries part of the repair. After the repair, masking the landmark position costs less
completion than on the undiversified policies (separated in two of three seeds), but not zero (\S\ref{sec:after}).

(4)~\textbf{Two measures misread this path, each on one task} (\S\ref{sec:position}): masking every
instruction word after the connector leaves t7 nearly intact, which, read alone, suggests the instruction's words go unused; a single-frame open-loop score misses t9's closed-loop gain from
transplanting all but the connector.

Hence we recommend judging robustness to rewording not by success rates alone but along the
path from language to action (\S\ref{sec:position}).

\section{Related work}\label{sec:related}
\paragraph{Paraphrase robustness and grounding gaps.}
Robustness of VLA policies to rewording has been measured suite-wide \citep{liberopara,industrialvla,liberoplus}
and across languages \citep{beyondenglish}, and vision-only ablations show how much of a suite needs no
language \citep{liberocf}. Suite-wide scores hide a failure's shape: \citet{retrieve} report the released
Spatial \model{} within five points of its original score under LIBERO-PRO's paraphrases \citep{liberopro},
and LIBERO-Para reads most failures as misidentifying the task. Yet with its $\pi_{0.5}$ policy's VLM frozen, some episodes fail near a successful trajectory, which LIBERO-Para suggests may be the VLM identifying the task while the
action-generating part fails to execute it \citep{liberopara}. Language is
not simply ignored: \citet{retrieve} explain failures on \model{} as instructions retrieving stored
trajectories, \citet{gsr} find a paraphrase's meaning kept inside the model, and \citet{encodednotcontrol}
find another named object linearly decodable inside \model{} while it does not control which object the
policy approaches, with a checkpoint trained on all of LIBERO by their account (Appendix~\ref{app:otherckpt}). The choice of object has been scored apart from success \citep{pi05,liberocf,robofollow,encodednotcontrol,retrieve};
under commonsense paraphrase, \citet{intact} find the intended object still chosen while success falls, the shape
of our claim~1. We add controls that tie the kept choice to the word's meaning (a nonce word in the same
slot, a synonym of the other landmark) and interventions inside the policy: \citet{intact} vary only the
inputs, and \citet{encodednotcontrol} vary the instruction and read internal states without changing them.

\paragraph{Intervention inside VLAs.}
Interventions on these models replace the feature sequence the action head reads with another task's, in \model{}
\citep{retrieve}, or a paraphrase's features with the original's, layer by layer \citep{gsr}, knock out attention \citep{vlatrace}, steer sparse
features or inject activations \citep{notallfeatures}, and attribute in closed loop \citep{impactvla}.
Concurrently, \citet{wulff} patch the diffusion transformer of a LIBERO-10 \model{} layer by layer in open loop,
find synonym replacement among the weakest perturbations, and call for closed-loop tests of it. We use
such established operations; what is new is their combination: a single word's token span, both sides
of one module, closed-loop scoring of choice and completion, and the same interventions after a repair.

\paragraph{Repair.}
Instruction diversification and augmentation \citep{stablelanguage}, counterfactual training that
makes language covary with the action \citep{cast}, inference-time guidance against a vision-only
prior \citep{liberocf,stablelanguage}, and re-binding the instruction through a new pathway with a
re-initialized action module \citep{gsr} repair fragility. Where a change lies has been located by
merging parameter groups, which places a VLA's robustness to scene changes in its language backbone \citep{retain}; by module swaps for a reinforcement-learning update \citep{lowrankrl}; and by finetuning one region at a time, which places adaptation to instruction shifts in the language
backbone across five architectures \citep{notalllayers}. With the VLM frozen, we instead test which
action-head tensors suffice to carry one repair, by transplant and by connector-only finetuning, and
re-apply the mask (and, in Appendix~\ref{app:flipft}, the paste) after the repair.

\section{Setup and calibration}\label{sec:setup}

\subsection{The readout path}\label{sec:path}
\model{} \citep{gr00tn1,gr00tn17card} runs a VLM cut from Cosmos-Reason2-2B \citep{cosmosreason2card}
(the parent VLM): its first 16 of 28 layers, further trained before the LIBERO finetuning
(Appendix~\ref{app:ckpt}). The policy reads the output of
layer 16 before the final normalization and passes it through the connector (\S\ref{sec:intro}) into a 32-block diffusion transformer, one vector per token: 128 image tokens
(two views), 22 tokens of the instruction's other words, template and separators (23 on t1, \S\ref{sec:calib}) and one
to four landmark tokens, 151--155 positions. The transformer denoises an action chunk
(8 of its action steps are executed before the next prediction) and cross-attends in alternate blocks to the instruction positions and to the image positions. The action head (\S\ref{sec:intro}) holds 537 tensors: 66
in the connector (a layer norm of 2 tensors and four self-attention blocks of 16 each) and 471 outside it (Appendix~\ref{app:subsets}). The VLM is causal and the images precede the instruction, so before the connector only the landmark position and
those after it (eight on the mirror pair) can differ between two instructions that differ only in the landmark word; after the connector
every position can. The released LIBERO-Spatial checkpoint was finetuned with the VLM frozen, as is
every policy we train (\S\ref{sec:repair}).

\subsection{Which tasks need the landmark, and which words can be tested}\label{sec:calib}
Across the ten tasks (t0--t9), removing the landmark phrase (leaving ``pick up the black bowl and place it on the plate'') lowers
completion to 20\% or less on four tasks---t1 (the bowl next to the ramekin), t4 (the bowl in the top drawer of the
wooden cabinet), t7 and t9: 7.8\%, 20.0\%, 0.0\% and 4.0\%; on t5 by 26
points (92\% to 66\%); and leaves the other five within 2 points (cell sizes: \S\ref{sec:stats}).
On these four tasks, the original instructions complete at 99.0\% and surface rewrites that keep the noun
(word order, verb, relation words) at 98--99\%.

Only the mirror pair (\S\ref{sec:intro}) can test a word's meaning against the other
landmark; t1 and t4 cannot, for opposite reasons. Task t1 counts against a semantic reading of its landmark: its five synonyms of ``ramekin'' complete at 12.8\% (32/250) while three nonce
words do at 70.9\% (107/151), unlike on the mirror pair (\S\ref{sec:reach}). We keep t1 for one check: whether pasting the
original word before the connector restores completion for this different failure too (\S\ref{sec:entry}). On t4 a filled slot, nonce word or synonym, completes above 90\% (139/150 and 239/250), so t4 cannot
show whether meaning is read.

\subsection{Interventions}\label{sec:interventions}
A \emph{paste} (\S\ref{sec:intro}) is made either before the connector, which then mixes the
pasted vectors, or after it, where the diffusion transformer reads them directly. Positions are pasted by
\emph{group}---the landmark, the rest of the instruction and the image positions (after the connector,
those that held image tokens)---singly or in pairs; the pair without the landmark is the \emph{complement}.
\emph{Flip} pastes the other landmark's vectors into an original-instruction episode and counts episodes
that approach the other bowl (approach: \S\ref{sec:stats}). \emph{Rescue} pastes the original
instruction's vectors into a synonym episode and counts completions. \emph{Masking} after the
connector removes the chosen positions from the action head's instruction cross-attention (their
vectors stay in place). Before the connector it does the same and also zeroes their vectors at the
connector's input, where the layer norm turns them into its bias, which still enters the
self-attention, so this mask also gives the connector an input it was not trained on. For the landmark
group, the \emph{donor} instruction's landmark vectors are averaged and pasted onto each position in the
\emph{recipient} episode's landmark span; positions after the landmark are paired by their offset from
its end. The connector has no positional embedding, so a vector copied to more positions weighs more in
its attention; pasting the original word's own averaged vector back into the original instruction
changes completion by an unresolved +3.0 points (Table~\ref{tab:subst_cells}). To check that an effect needs the landmark position,
not merely a paste, position controls paste at as many positions outside the landmark as its span has (\emph{matched} positions),
averaged the same way, or at the verb's one position.
Before the connector a control paste can carry the landmark word only at positions after the landmark
(\S\ref{sec:path}); of the matched positions, one on t7 and none on t9 lie there, so those
controls mostly test the averaged paste itself. Each paste and mask
is logged and checked (Appendix~\ref{app:receipts}).

\subsection{Metrics and statistics}\label{sec:stats}
In closed loop the policy's actions are executed and each prediction uses
the resulting observation. An episode \emph{approaches} a bowl if the gripper comes within 10~cm
of it (Appendix~\ref{app:thr}), and \emph{completes} if LIBERO's success check for the task is
met---the success a suite reports. In open loop (used only in \S\ref{sec:position}) the policy
predicts an action chunk for a recorded frame and nothing is executed. Each cell (one instruction, task, condition and policy) is 50
closed-loop episodes (five parallel environments; 51 to 53 when the last batch
overruns). A \emph{run} is one batch of cells measured together; conditions within a run share the environment seed (0 unless stated), applied at
the first reset of each parallel environment. Under that seed initial placements are drawn
from the task's placement regions, not from LIBERO's fixed initial-state files; conditions are
not paired episode by episode, since which placements fill a cell depends on how long episodes last.

We compare cells as independent samples by
a pooled two-proportion $z$ with the threshold $|z|\ge 3$, fixed before measurement (about $p<0.003$
two-sided; roughly 30 points for two 50-episode cells near one half). Differences
reaching it are \emph{separated}; smaller ones are reported as \emph{unresolved}, never as equality.
Fisher's exact test at the same level agrees on every two-cell comparison. We compare
finetuned policies per training seed with the same $z$ (with the unpooled variance for differences of two
masking costs). Within a seed, rates and point differences are computed from the pooled completions
of the compared cells, the counts the $z$ uses; we report training seeds separately, never pooled. \emph{Decision cells} are
the synonym cells whose unmodified completion was at most 40/50 in run 1, which measured all
ten synonyms: hotplate on t7; dresser, chest of drawers, sideboard and bureau on t9 (Table~\ref{tab:subst_cells}). Later runs measure only
these five and keep those still at most 40/50 in that run (all five, on the released policy with environment seed 0). The other
five synonyms, already completed at 90--100\%, leave little room to
show a rescue. Repeated runs of the same cells fall within binomial scatter (released policy, five decision
cells, runs 1--3: 158, 149 and 154 of 250; $|z|\le0.9$). With the baseline taken from another run, the rescue still gains and the mask still costs
(Appendix~\ref{app:bcells}). The repeats detect no
run effect but do not prove the independence assumed across runs.
Stratifying by cell changes no verdict; of the 99 comparisons listed in Appendix~\ref{app:stats}, three separations, each in one
seed, fail Holm's correction.

\section{Where meaning reaches, and where choice and completion part}\label{sec:locus}

\subsection{How far the synonym reaches}\label{sec:reach}
\begin{table}[t]
  \centering\footnotesize
  \caption{How far an unpresented landmark word reaches on the mirror pair (released policy; two runs separate from
  runs 1--3 of \S\ref{sec:stats}, the second giving the nonce and other-landmark-synonym rows; episodes approaching each bowl, and episodes completing the original instruction's task; $n$: episodes, pooled over words; target:
  the bowl the original instruction names, other: the pair's other bowl).}
  \label{tab:reach}
  \begin{tabular}{@{}llrrrr@{}}
\toprule
Task & Instruction & $n$ & To target & To other bowl & Completed \\
\midrule
t7 & Original instruction & 50 & 50 & 0 & 49 \\
t7 & Synonym of the landmark (5 words) & 250 & 237 & 3 & 225 \\
t7 & Nonce word in the landmark slot (3 words) & 150 & 0 & 9 & 0 \\
t7 & The other landmark's word & 50 & 0 & 50 & 0 \\
t7 & Unpresented synonym of the other landmark & 50 & 0 & 41 & 0 \\
t7 & Landmark phrase removed & 50 & 0 & 4 & 0 \\
\midrule
t9 & Original instruction & 50 & 50 & 0 & 50 \\
t9 & Synonym of the landmark (5 words) & 250 & 223 & 0 & 169 \\
t9 & Nonce word in the landmark slot (3 words) & 150 & 4 & 0 & 2 \\
t9 & The other landmark's word & 50 & 0 & 50 & 0 \\
t9 & Unpresented synonym of the other landmark & 50 & 0 & 50 & 0 \\
t9 & Landmark phrase removed & 50 & 5 & 0 & 2 \\
\bottomrule
\end{tabular}

\end{table}
Table~\ref{tab:reach} shows that (a)--(c) of \S\ref{sec:intro} hold, judged by the $z$ of \S\ref{sec:stats} on approach: to the
named bowl, synonyms against nonce words; to the other bowl, the other landmark's synonym against
the landmark's own synonyms (every $z\ge14.7$). The five synonyms send
the policy to the named bowl in 237/250 episodes on t7 and 223/250 on t9, and the nonce words in 0/150 and 4/150.
An unpresented synonym of the other landmark (sideboard on t7, hob on t9) sends it to the other bowl
in 41/50 and 50/50 (the other landmark's training word: 50/50 on both); with the landmark phrase
removed, the policy almost never approaches either bowl.
Nonce words still come within 15~cm of the cabinet's bowl in about half the episodes on both tasks, so what they lack is
the choice (Appendix~\ref{app:thr}). At 6, 10 and 15~cm alike, synonyms rarely approach the other bowl
while hob does (Table~\ref{tab:thresholds}).
As an external control on whether the
wording is resolvable, we ask the 28-layer parent VLM with its language-model head which of
the two bowls the instruction names. Over the four landmark-dependent tasks it
scores the correct bowl higher in 199/200 cases (t4's 50 cases do not test the synonym; Appendix~\ref{app:pvlm}); this is not a readout of the policy's own VLM (\S\ref{sec:path}).

Completion is where synonyms fall short: 67.6\% on t9 (27--46/50 per word) against the training word's 100\%
($z=-4.7$); on t7 the pooled 90.0\% against 98\% is unresolved ($z=-1.84$); but one word falls short
(hotplate, 31/50 against 49/50, $z=-4.5$; the other four 45--50/50). Of the 81 failed t9 synonym episodes, 54 approach the
target, 27 neither bowl and none the other bowl. In t4's slot the same five t9 words complete 239/250 (\S\ref{sec:calib}), so the shortfall is not a property of the word alone.

\begin{table}[t]
  \centering\scriptsize
  \caption{Flip, rescue and masking (\S\ref{sec:interventions}) before and after the connector (released policy, mirror pair unless
  stated). B: differences (points, two-proportion $z$) from the unmodified synonym cells of the same run (run 1 or 2; seed-100 and t1 rows: own runs).
  C: run 3 (seed-100 row: own run). Instruction except landmark is the
  rest of the instruction; all except landmark, the complement. $^{\dagger}$Only positions after the landmark can
  differ before the connector (\S\ref{sec:path}).}
  \label{tab:locus}
  \begin{tabular}{@{}llcc@{}}
\toprule
\multicolumn{4}{@{}l}{\textbf{A. Flip: episodes approaching the other bowl (of 50)}} \\
Site & Positions replaced & t7 & t9 \\
\midrule
before connector & landmark & 50 & 50 \\
before connector & instruction except landmark$^{\dagger}$ & 0 & 0 \\
before connector & all except landmark$^{\dagger}$ & 0 & 0 \\
\midrule
after connector & landmark & 0 & 1 \\
after connector & instruction except landmark & 0 & 0 \\
after connector & image & 6 & 1 \\
after connector & landmark + rest of instruction & 41 & 47 \\
after connector & landmark + image & 50 & 50 \\
after connector & rest of instruction + image (all except landmark) & 49 & 42 \\
after connector & all & 50 & 50 \\
\midrule
\multicolumn{4}{@{}l}{\textbf{B. Rescue: completed, five decision cells pooled}} \\
Site & Positions replaced & \multicolumn{2}{l}{Completed (difference, $z$)} \\
\midrule
unmodified & -- & \multicolumn{2}{l}{158/250 (run 1), 149/250 (run 2)} \\
before connector (run 1) & landmark & \multicolumn{2}{l}{244/250 (+34.4, $z=9.69$)} \\
after connector (run 1) & landmark & \multicolumn{2}{l}{217/250 (+23.6, $z=6.09$)} \\
after connector (run 2) & instruction except landmark & \multicolumn{2}{l}{231/251 (+32.4, $z=8.48$)} \\
after connector (run 2) & image & \multicolumn{2}{l}{215/250 (+26.4, $z=6.63$)} \\
after connector (run 2) & all except landmark & \multicolumn{2}{l}{246/250 (+38.8, $z=10.65$)} \\
before connector, env.\ seed 100 & landmark & \multicolumn{2}{l}{+35.5, $z=9.00$ (4 cells)} \\
after connector, env.\ seed 100 & all except landmark & \multicolumn{2}{l}{+34.0, $z=8.52$ (4 cells)} \\
\midrule
\multicolumn{4}{@{}l}{\textit{Task t1 (five synonyms of ``ramekin''; not a mirror pair)}} \\
unmodified & -- & \multicolumn{2}{l}{29/254} \\
before connector & landmark & \multicolumn{2}{l}{251/251 (+88.6, $z=20.03$)} \\
before connector & matched positions elsewhere & \multicolumn{2}{l}{8/101 vs.\ 7/102 (2 words; +1.1, $z=0.29$)} \\
after connector & all except landmark & \multicolumn{2}{l}{187/250 (+63.4, $z=14.38$)} \\
\midrule
\multicolumn{4}{@{}l}{\textbf{C. Masking the landmark position after the connector: completed}} \\
Instruction & Policy & \multicolumn{2}{l}{Masked vs.\ unmasked (difference, $z$)} \\
\midrule
synonym, 5 decision cells & released & \multicolumn{2}{l}{98/250 vs.\ 154/250 ($-$22.4, $z=-5.01$)} \\
synonym, env.\ seed 100 & released & \multicolumn{2}{l}{$-$20.3, $z=-3.52$ (3 cells)} \\
original, t7 / t9 & released & \multicolumn{2}{l}{49/50 vs.\ 49/50 (+0.0); 47/50 vs.\ 49/50 ($-$4.0)} \\
\bottomrule
\end{tabular}

\end{table}

\subsection{Before the connector the landmark position alone suffices to set the choice}\label{sec:preneck}
Pasting the vector of the other landmark word
before the connector at the landmark position alone---one or two tokens---flips 50/50 episodes on
both tasks (Table~\ref{tab:locus}A; 49/50 and 50/50 with environment seed 100, Appendix~\ref{app:ledger}). The eight positions after the landmark (\S\ref{sec:path}) do differ
between the two instructions, yet pasting them with the rest of the instruction flips none. The absence of flips does not imply that these positions lack landmark information.

\subsection{After the connector the choice is distributed}\label{sec:postneck}
After the connector a single group rarely moves the choice: the same paste at the landmark
position flips 0/50 on t7 and 1/50 on t9, the rest of the instruction 0 and 0, and the image
group 6 and 1, though it costs t7 completion (Appendix~\ref{app:bcells}). Any two groups flip the choice (41--50 of 50; Table~\ref{tab:locus}A), as do all
positions. The groups differ in size (\S\ref{sec:path}) and the pairs have no matched-size control, so
the result concerns these groups, not how many positions suffice.

\subsection{A synonym's failure is recoverable on both sides of the connector}\label{sec:entry}
A synonym's failure is recoverable at the landmark position \emph{before} the connector---a
sufficient point of intervention, which does not by itself show where the failure originates. Pasting the original
instruction's vector there restores completion in the five decision
cells from 158/250 to 244/250 (Table~\ref{tab:locus}B; +34.4 points, $z=9.69$; every cell 47--50/50), and also with
environment seed 100.

After the connector the same paste restores less: +23.6 points, 10.8 below the pre-connector
rescue ($z=-4.5$). The complement restores +38.8, the rest of the instruction +32.4 and the image group
+26.4 (Table~\ref{tab:locus}B, run 2): a rescue there needs no paste at the landmark position, and single groups that barely move the
choice (\S\ref{sec:postneck}) still restore completion. On t1 the pre-connector rescue takes the five synonyms from
29/254 to 251/251, 25.2 points above the post-connector complement ($z=8.51$); a matched
position control (\S\ref{sec:interventions}) gains an unresolved +1.1.

\subsection{After the connector the landmark position still matters}\label{sec:residual}
Masking the landmark position after the connector (\S\ref{sec:interventions}), where pasting it alone
barely moves the choice (\S\ref{sec:postneck}), still costs the synonym decision cells 22.4 points of
completion ($z=-5.01$; Table~\ref{tab:locus}C) and 20.0 points of approach to the
target (212/250 to 162/250, $z=-5.15$); with environment seed 100, 20.3 of completion. On two cells, masking matched positions (\S\ref{sec:interventions}) moves completion by at most three episodes (Table~\ref{tab:ko_cells}). On the original
instruction the cost is unresolved (t7 49/50 vs.\ 49/50, $z=0.0$; t9 47/50 vs.\ 49/50, $z=-1.02$).

\section{Repair, and which tensors suffice to carry it}\label{sec:repair}

\subsection{Diversifying the landmark word repairs completion on t9}\label{sec:var}
We finetune the released checkpoint on the same demonstrations (recipe in Appendix~\ref{app:recipe})
under three instruction sets. The \emph{undiversified} policy keeps the original ten instructions. The
\emph{landmark-diversified} policy (\emph{diversified} for short) draws the landmark word of
three tasks at random from a list per task (t1: ramekin, custard cup, small dish, little pot;
t7: stove, hotplate; t9: wooden cabinet, chest of drawers). The \emph{verb-diversified} policy instead draws
the verb phrase of the same three tasks, from lists of the same sizes.

\begin{table}[htb]
  \centering\scriptsize
  \caption{Repair by finetuning, by transplant and by connector-only finetuning, and the post-connector mask after the repair (three training seeds;
  environment seed 0). A--B: unpresented synonyms are, for every policy, the
  eight mirror-pair synonym cells outside the landmark lists of \S\ref{sec:var}; four are on t9 (A: mean of the four words' rates). C: the masking of
  Table~\ref{tab:locus}C on the undiversified and landmark-diversified policies. In C, all rows use the cells the rule of
  \S\ref{sec:stats} selects for that seed's undiversified policy; the rule drops training seed 101's
  hotplate cell and would select none on the diversified policies (all five cells fixed: Appendix~\ref{app:maskfixed}).}
  \label{tab:repair}
  \begin{tabular}{@{}lccc@{}}
\toprule
Policy & seed 101 & seed 202 & seed 303 \\
\midrule
\multicolumn{4}{@{}l}{\textbf{A. Completed, task t9, four unpresented synonyms (\%)}} \\
finetuned, undiversified & 63.0 & 61.5 & 65.0 \\
finetuned, verb-diversified & 69.5 & 66.0 & 64.7 \\
finetuned, landmark-diversified & 99.0 & 98.5 & 100.0 \\
undiversified + 66 connector tensors of the diversified policy (connector transplant) & 98.5 & 98.0 & 98.0 \\
undiversified + the diversified policy's other 471 tensors (all-but-connector transplant) & 87.5 & 91.0 & 88.0 \\
connector-only finetuning, undiversified & 59.5 & 61.0 & 65.0 \\
connector-only finetuning, landmark-diversified & 99.0 & 98.5 & 98.5 \\
\midrule
\multicolumn{4}{@{}l}{\textbf{B. Unpresented synonyms, mirror pair: difference in points ($z$)}} \\
landmark-diversified $-$ undiversified & +18.0 (8.4) & +20.2 (9.2) & +17.8 (8.6) \\
verb-diversified $-$ undiversified & +3.0 (1.1) & +1.9 (0.7) & $-$0.7 ($-$0.3) \\
connector transplant $-$ undiversified & +17.8 (8.2) & +20.0 (9.0) & +17.3 (8.3) \\
landmark-diversified $-$ connector transplant & +0.2 (0.3) & +0.3 (0.4) & +0.5 (0.8) \\
landmark-diversified $-$ all-but-connector transplant & +5.0 (3.7) & +4.8 (3.9) & +5.5 (4.4) \\
all-but-connector transplant $-$ undiversified & +13.0 (5.5) & +15.4 (6.5) & +12.3 (5.3) \\
connector-only finetuning: landmark-diversified $-$ undiversified & +20.5 (9.0) & +19.8 (8.9) & +17.2 (8.1) \\
\midrule
\multicolumn{4}{@{}l}{\textbf{C. Masking cost after the connector, synonym decision cells: points ($z$)}} \\
finetuned, undiversified & $-$19.0 ($-$3.8) & $-$26.2 ($-$6.1) & $-$27.2 ($-$6.2) \\
finetuned, landmark-diversified & $-$8.0 ($-$3.7) & $-$6.4 ($-$3.5) & $-$5.6 ($-$3.2) \\
difference, diversified $-$ undiversified & +11.0 (2.1) & +19.8 (4.4) & +21.6 (4.7) \\
\bottomrule
\end{tabular}

\end{table}
Relative to the undiversified policy, landmark diversification raises completion on the mirror pair's
eight unpresented synonym cells by 18.0, 20.2 and 17.8 points in the three training seeds ($z=8.4$ to 9.2;
Table~\ref{tab:repair}A--B), and verb diversification by 3.0, 1.9 and $-$0.7, unresolved in every
seed ($|z|<1.2$). The effect is on t9 (61.5--65.0\% to 98.5--100\%); t7 is at ceiling for
every policy (96.0--100\%).

One unfavorable property goes with the repair:
when the landmark cannot be read (nonce word or landmark phrase removed;
Table~\ref{tab:bias}), the diversified policy, unlike the undiversified one, drifts to the bowl on the cabinet,
t9's target, so on t9 drift and repair point to the same bowl. With a nonce word it completes
t9 in only 12, 10 and 15 of 50, so drift alone does not explain the repair.

\subsection{The connector's 66 tensors suffice for the repair; other tensors carry part of it}\label{sec:transplant}
The \emph{connector} transplant, which copies the 66 connector tensors of the landmark-diversified
policy into the undiversified policy of the same seed without training, carries the repair: on the
unpresented synonyms it adds 17.8, 20.0 and 17.3 points to the undiversified policy ($z\ge8.2$). The completion-rate differences (fully finetuned diversified policy minus transplant) are +0.2, +0.3 and +0.5 points, unresolved
in every seed ($|z|<0.9$; Table~\ref{tab:repair}B; 95\% intervals within $-1.6$ to $+2.1$, Appendix~\ref{app:stats}). The
\emph{all-but-connector} transplant, which copies the other 471 action-head tensors instead,
carries about seven tenths of the diversified policy's gain on the mirror pair (0.72, 0.76, 0.69), so other tensors also carry part of the repair.
In every seed the diversified policy exceeds this transplant; the margin is what the diversified
connector adds to the copied tensors.
None of the four subsets of these 471 tensors, transplanted alone, carries the 471's share (Appendix~\ref{app:subsets}).

Connector-only finetuning (last rows of Table~\ref{tab:repair}A and B) shows the same sufficiency by training instead of
copying: on the landmark-diversified instructions it exceeds connector-only finetuning on the original instructions
by 17 to 21 points on the unpresented synonyms. On the five decision cells (fixed as in Appendix~\ref{app:maskfixed}),
connector-only minus full finetuning is unresolved in every seed with either instruction set
($|z|\le2.0$; Appendix~\ref{app:stats}).

Drift alone does not explain these repairs either: with a nonce word, the diversified connector-only policy completes t9
in 4/51, 6/50 and 6/51 and the connector transplant in 6/50, 2/50 and 3/51, against 98.0--99.0\% with
the unpresented synonyms (Table~\ref{tab:repair}A; their drift: Appendix~\ref{app:drift}).

\subsection{After the repair, masking the landmark position still costs completion, but less}\label{sec:after}
On the landmark-diversified policies the post-connector mask of \S\ref{sec:residual} costs less
completion than on the undiversified ones (Table~\ref{tab:repair}C, whose cells include words the diversified policies trained on): 0.42, 0.24 and 0.21 times as much. The diversified policies' cost itself is separated from zero in each seed (seed 303's not after
Holm's correction; Appendix~\ref{app:stats}). The difference between the two costs separates in two seeds and is unresolved in
seed 101; with the five cells fixed it separates in all three (Appendix~\ref{app:maskfixed}). On the three t9 decision cells with unpresented synonyms,
the ratio is 0.41, 0.48
and 0.24, and the difference separates in one seed ($z=2.1$, $2.0$ and $4.3$). There the smaller cost is not
explained by the drift of \S\ref{sec:var} alone: the masked diversified policies
still complete 134, 132 and 137 of 150, far above their nonce-word completions (Table~\ref{tab:bias}; hotplate: Appendix~\ref{app:maskfixed}). The cost also remains on those three cells: they lose 9.3, 11.3 and 8.0
points ($z=-3.42$, $-4.03$ and $-3.28$).

\section{Discussion and conclusion}\label{sec:position}
\begin{figure}[t]
  \centering\footnotesize
  \includegraphics[width=\linewidth]{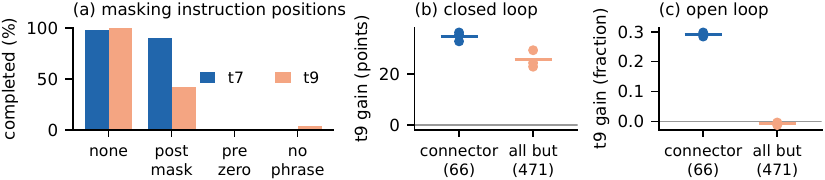}
  \caption{Two measures that misread the path. (a)~Released policy: completion unmasked, with every
  instruction-word position masked after the connector or also zeroed at its input
  (\S\ref{sec:interventions}), and with the landmark phrase removed. (b)~Closed-loop gain on t9's four
  unpresented synonyms from the connector and all-but-connector transplants into the undiversified
  policy (dots: three seeds, some overlapping; colors here mark the transplant, not the task as in (a); lines: mean). (c)~The same in the open-loop retained fraction, which
  averages all five t9 synonyms, one a finetuning word (Appendix~\ref{app:openloop}); units differ from (b).}
  \label{fig:misread}
\end{figure}

\paragraph{Two measures that misread the path.}
The first measure is the post-connector mask on every instruction-word position with the original
instruction (template and separator tokens kept): it leaves t7 at 45/50 against 49/50 unmasked ($z=-1.7$, unresolved;
Fig.~\ref{fig:misread}a). Read alone, this suggests that the action head does not use the instruction's words. It does not
show that: on t9 the same mask gives 21/50 against 50/50 ($z=-6.4$), and on t7 removing the landmark phrase gives
0/50, as does the pre-connector mask on both tasks, though it also feeds the
connector an untrained input (\S\ref{sec:interventions}).

The second measure, the open-loop retained fraction, scores a synonym's action chunk on single starting
frames, robot state zeroed (Appendix~\ref{app:openloop}): the original instruction's chunk scores 1, the other
landmark word's 0. The all-but-connector transplant changes t9's score by $-0.006$, $-0.012$
and $-0.003$ by seed (paired $|t|\le1.6$ over ten frames, unresolved), while the connector
transplant adds $+0.284$ to $+0.298$; this score, read alone, wrongly suggests that only the connector carries
the repair. In closed loop the all-but-connector transplant yields +24.5, +29.5 and +23.0
points on t9 (Fig.~\ref{fig:misread}b,c), and its gain on the mirror pair separates in every seed
(Table~\ref{tab:repair}B), a divergence of open- and closed-loop scores also
reported elsewhere \citep{impactvla,closedloopcollapse}.

\paragraph{What we could not determine.}
On the failure: whether what breaks completion lies in the landmark-position vector, as we suspect from
the pre-connector rescue, and what predicts which synonyms fall short (the
parent VLM's choice, correct in nearly every case, does not; \S\ref{sec:reach}). On the connector: why two groups suffice after it
(\S\ref{sec:postneck}). On the repair: what the masking cost that remains after the repair (\S\ref{sec:after}) reflects; whether the repair holds for a
word naming the bowl opposite the drift of \S\ref{sec:var}; and whether training only the
other 471 tensors repairs as far as full finetuning does.

\paragraph{Limitations.}
Our claims concern only the mirror pair, where the landmark word chooses the bowl; evidence
that these policies lean on vision \citep{liberoplus,liberocf,encodednotcontrol} (\S\ref{sec:calib})
stands. The claims rest on one checkpoint of one architecture; five decision cells, four
of them furniture synonyms on t9; one word per task for criterion (c); and English landmark words
only. On the released LIBERO-Object checkpoint, a synonym sharing no token with the trained object noun
fails to complete (1/50; Appendix~\ref{app:otherckpt}); with its approach unrecorded, whether the pattern holds beyond Spatial is open. The repair acts on t9
(\S\ref{sec:var}) and was measured with environment seed 0 only.
Decision cells change with the environment seed (Table~\ref{tab:locus}B, C; Appendix~\ref{app:ledger}).
The rescue's sensitivity to how many
positions the averaged vector is copied to is untested.
We do not decompose the connector below its 66 tensors, and we test one open-loop measure, on
starting frames with the state zeroed (Appendix~\ref{app:openloop}).

\paragraph{Summary and recommendation.}
On the mirror pair of this released VLA, unpresented landmark synonyms reach the choice of bowl,
yet some fall short of completion (claim~1). That shape has been reported before (\S\ref{sec:related});
this case study adds controls that tie the choice to the word's meaning, and interventions on
one word's tokens before and after the connector in closed loop. The interventions locate where the action gap can be changed: the
landmark word acts differently on the connector's two sides, and training only the connector's tensors on diversified instructions, or copying them from the landmark-diversified policy, repairs
completion on t9, though other tensors carry part of the repair (claims~2--3). Since success rates alone hide the split of claim~1 and two single measures misread this path (claim~4), we recommend that, for now, robustness to rewording be judged in
closed loop, intervening on both sides of the connector, scoring the choice and completion separately.

\clearpage
\acknowledgments{We would like to thank the members of the AI Robotics Business Division and the AI Technical Sector for their advice and comments on this paper. We acknowledge the support of MAMEZO Co., Ltd., which provided the research resources for this work, including the computing environment for our measurements. Finally, we thank the organizers of the CoRL 2026 Workshop on The Action Gap (TAG 2026) for the opportunity to present this work, and the area chair and the reviewers for their constructive feedback.}

\paragraph{Use of AI assistants.}
ChatGPT, Codex and Claude Code were used to assist with literature review, code development, analysis
review, and manuscript drafting and editing. The authors designed the experiments, verified the reported
results and references, reviewed the implementation, and determined all research claims and conclusions.
The authors take full responsibility for the content.

\bibliography{refs}

\clearpage
\appendix
\section{Experiments behind the claims}\label{app:ledger}
Table~\ref{tab:ledger} lists the experiments the main text and Appendix~\ref{app:obs} rest on.
{\footnotesize
\setlength{\tabcolsep}{4pt}
\begin{longtable}{@{}>{\raggedright\arraybackslash}p{0.27\linewidth}>{\raggedright\arraybackslash}p{0.30\linewidth}>{\raggedright\arraybackslash}p{0.37\linewidth}@{}}
\caption{The experiments behind the claims: what each compares and what it gave. Three numbers
separated by slashes are training seeds 101 / 202 / 303; differences are in points; ``unresolved''
is $|z|<3$ (\S\ref{sec:stats}). In the released-policy rows, two numbers separated by a comma are t7, t9;
$x/n$ is $x$ of $n$ episodes.}
\label{tab:ledger}\\
\toprule
Experiment (section) & Comparison & Result \\
\midrule
\endfirsthead
\multicolumn{3}{@{}l}{\emph{Table~\ref{tab:ledger}, continued}}\\
\toprule
Experiment (section) & Comparison & Result \\
\midrule
\endhead
\midrule
\endfoot
\bottomrule
\endlastfoot
\multicolumn{3}{@{}l}{\emph{Released policy}} \\
calibration with the landmark phrase removed (\S\ref{sec:calib}) & each task with the landmark phrase removed & t1 7.8\%, t4 20.0\%, t7 0.0\%, t9 4.0\%; the other six 66--100\% \\ \addlinespace[3pt]
surface rewrites that keep the noun (\S\ref{sec:calib}) & the original instruction vs.\ rewrites of word order and verb, or of relation words, on t1, t4, t7 and t9 & 99.0\% vs.\ 98.0--99.0\% (mean over the four tasks) \\ \addlinespace[3pt]
synonyms and nonce words on t1 and t4 (\S\ref{sec:calib}) & completion, synonyms vs.\ nonce words & t1: synonyms 12.8\% (32/250) vs.\ nonce 70.9\% (107/151); t4: synonyms 239/250, nonce 139/150 \\ \addlinespace[3pt]
synonyms, nonce words and the other landmark's synonym (\S\ref{sec:reach}) & approach to each bowl (criteria (a)--(c)) and completion against the training word (d), mirror pair & named bowl 237/250, 223/250 vs.\ nonce 0/150, 4/150; other bowl 41/50, 50/50 vs.\ the landmark's own synonyms 3/250, 0/250 (every $z\ge14.7$); completion 225/250 vs.\ 49/50 (unresolved), 169/250 vs.\ 50/50 ($z=-4.7$); hotplate 31/50 ($z=-4.5$) \\ \addlinespace[3pt]
parent-VLM control (\S\ref{sec:reach}, Appendix~\ref{app:pvlm}) & which bowl the edited instruction names, by summed log-likelihood, on t1, t4, t7 and t9 & correct in 199/200 synonym cases (ten frames per word; t4's two candidates share the noun and differ only in the relation phrase, so its 50 cases do not test the synonym) \\ \addlinespace[3pt]
paste at the landmark position (\S\ref{sec:preneck}, \S\ref{sec:entry}; run 1 of Table~\ref{tab:locus}) & flip and rescue, before vs.\ after the connector; position controls; the original word's own vector pasted & flip 50/50 before, 0--1/50 after; rescue +34.4 before, +23.6 after (before $-$ after 10.8); controls unresolved ($|z|<1.2$): matched positions $-$2 before the connector (where they can carry the original landmark word at one position on t7 and none on t9, so they control for the averaged paste itself) and verb +2 (one position, left unchanged there by the paste), verb $-$4 and matched +8 after; own vector pasted +3.0 \\ \addlinespace[3pt]
position groups (\S\ref{sec:postneck}, \S\ref{sec:entry}; run 2 of Table~\ref{tab:locus}) & flip and rescue by group after the connector; the two groups tested before it & after: all positions 50, 50; complement 49, 42; rest of the instruction 0, 0; image 6, 1; rescue from complement +38.8. Before: rest of the instruction 0, 0; image vectors identical (max.\ difference 0.0) \\ \addlinespace[3pt]
landmark plus one other group (\S\ref{sec:postneck}; a separate run, so Table~\ref{tab:locus}A draws on three runs) & flip after the connector & landmark + rest of instruction 41, 47; landmark + image 50, 50 \\ \addlinespace[3pt]
post-connector masking of the landmark position (\S\ref{sec:residual}) & synonym decision cells and original instruction, masked vs.\ not; matched other positions & synonyms $-$22.4 ($z=-5.01$); original 0.0 and $-$4.0, both unresolved; other positions unresolved \\ \addlinespace[3pt]
masking every instruction word (\S\ref{sec:position}) & original instruction, every instruction-word position masked after the connector, or also zeroed at its input, vs.\ unmasked & after: 45/50 vs.\ 49/50 ($z=-1.7$, unresolved), 21/50 vs.\ 50/50 ($z=-6.4$); also zeroed: 0/50, 0/50 \\ \addlinespace[3pt]
environment seed 100 (\S\ref{sec:entry}, \S\ref{sec:residual}) & the flips, both rescues and the mask repeated & flips at the landmark 49, 50 before and 0, 4 after the connector, complement after 47, 45; pre-connector rescue +35.5; post-connector complement +34.0; mask $-$20.3; the decision cells change (pastes: bureau out at 43/50, four cells; mask: dresser and bureau out at 42/50, three cells) \\ \addlinespace[3pt]
task t1 (\S\ref{sec:entry}) & rescue before the connector, complement after, position control & +88.6; +63.4 (25.2 short of the pre-connector rescue); control including positions after the landmark +1.1, unresolved \\ \addlinespace[3pt]
\multicolumn{3}{@{}l}{\emph{Finetuned policies and transplants (three training seeds)}} \\
landmark diversification (\S\ref{sec:var}) & diversified vs.\ undiversified and vs.\ verb-diversified, unpresented mirror-pair synonyms & +18.0 / +20.2 / +17.8 over undiversified; +15.0 / +18.3 / +18.5 over verb-diversified \\ \addlinespace[3pt]
control with the landmark phrase removed after diversification (Appendix~\ref{app:drift}) & diversified vs.\ undiversified, landmark phrase removed & t9 10 / 7 / 7 of 50 vs.\ 1 / 0 / 0 (each seed unresolved, $z\le 2.88$); t1 33 / 36 / 32 of 50 vs.\ 7 / 7 / 5 \\ \addlinespace[3pt]
connector transplant (\S\ref{sec:transplant}) & vs.\ undiversified, unpresented mirror-pair synonyms; the diversified policy vs.\ the transplant & +17.8 / +20.0 / +17.3 ($z=8.2$ / 9.0 / 8.3); +0.2 / +0.3 / +0.5 ($z=0.3$ / 0.4 / 0.8), unresolved \\ \addlinespace[3pt]
all-but-connector transplant (\S\ref{sec:transplant}) & diversified vs.\ the transplant; the transplant vs.\ undiversified & +5.0 / +4.8 / +5.5 ($z=3.7$ / 3.9 / 4.4); +13.0 / +15.4 / +12.3 ($z=5.5$ / 6.5 / 5.3) \\ \addlinespace[3pt]
connector transplant with the landmark phrase removed (Appendix~\ref{app:drift}) & completion on t7 and t9 with the landmark phrase removed & 0--2 completed of 50--51 \\ \addlinespace[3pt]
subsets of the other 471 tensors (Appendix~\ref{app:subsets}) & each subset transplanted alone vs.\ undiversified & instruction cross-attention +6.5 / +6.2 / +4.2; image +6.0 / +7.5 / +1.3; self-attention +5.0 / +3.5 / +3.3; input and output tensors +3.3 / $-$1.0 / $-$0.8; none separates from the undiversified policy, each falls short of the 471 together in every seed ($z\le-3.1$) \\ \addlinespace[3pt]
open-loop measure of the transplants (\S\ref{sec:position}, Appendix~\ref{app:openloop}) & retained fraction of the action chunk, transplant $-$ undiversified; mirror pair pooled, and t9 alone & an open-loop metric, not completion; values in \S\ref{sec:position} and Appendix~\ref{app:openloop} \\ \addlinespace[3pt]
masking on the finetuned policies (\S\ref{sec:after}) & post-connector mask of the landmark position, undiversified vs.\ diversified policies & $-$19.0 / $-$26.2 / $-$27.2 vs.\ $-$8.0 / $-$6.4 / $-$5.6 on the same cells (four in seed 101; diversified separated from zero in each seed); difference +11.0 ($z=2.1$, unresolved) / +19.8 (4.4) / +21.6 (4.7); five cells fixed +18.0 / +19.8 / +21.6 \\ \addlinespace[3pt]
flips on the finetuned policies (Appendix~\ref{app:flipft}) & three of the flips of Table~\ref{tab:locus}A on t7 and t9 & t9: undiversified policies complete 27 / 37 / 39 of 50--51 with the other landmark's vector at the post-connector landmark position, against 50/50 unmodified in a separate run (no same-run baseline, so no verdict); diversified policies: 15 / 6 / 11 of 50 approach the other bowl \\ \addlinespace[3pt]
connector-only finetuning (\S\ref{sec:transplant}) & the 66 connector tensors trained alone on the landmark-diversified vs.\ the original instructions; each vs.\ the fully finetuned policy on the same instructions & diversified $-$ undiversified: unpresented mirror-pair synonyms +20.5 / +19.8 / +17.2 ($z=9.0$ / 8.9 / 8.1), five decision cells +48.8 / +48.4 / +41.6. On the same five cells vs.\ full finetuning: $-$0.4 / 0.0 / $-$0.8 (diversified), $-$8.8 / $-$0.2 / +2.0 (undiversified), all unresolved. Original instruction at least 48/50 in all twelve cells; diversified on t9, nonce word 4 / 6 / 6 of 50--51, landmark phrase removed 2 / 4 / 1 of 50--53 \\ \addlinespace[3pt]
\multicolumn{3}{@{}l}{\emph{Other checkpoints, and finetuning on Goal demonstrations}} \\
other checkpoints and covarying destination phrases (Appendix~\ref{app:otherckpt}) & released Goal, Spatial and Object checkpoints with other destination or object words; Spatial finetuned on Goal demonstrations whose destination phrase covaries with the action in $P$\% of the three bowl tasks' demonstrations & Goal: named fixture 297/300 off the diagonal, destination synonyms 248/250 vs.\ 54/150 with the phrase removed; Spatial: another destination not followed (196/200 on the plate); Object: another object named 1/500, a synonym sharing no token 1/50; destination following 0.16--0.19 at $P=0$ to 0.90--0.95 at $P=100$, shuffled phrases 0.32 \\ \addlinespace[3pt]
\end{longtable}}

\section{Procedures and decision cells}\label{app:cells}
\begin{figure}[h]
  \centering
  \includegraphics[width=0.5\linewidth]{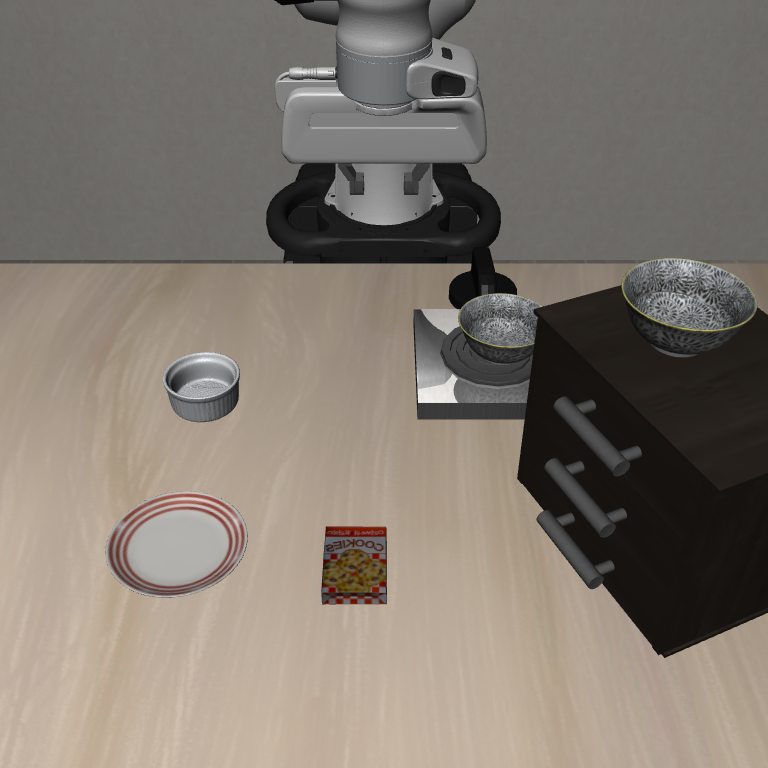}
  \caption{The scene of t7 and t9 (environment seed 0; the policy's third-person camera view, rendered at
  768 pixels). t7 names the bowl on the stove (center right) and t9 the bowl on the wooden cabinet (far
  right); both place it on the plate (left).}
  \label{fig:scene}
\end{figure}

Figure~\ref{fig:scene} shows the scene that the mirror pair (t7 and t9) shares.

\subsection{Approach threshold}\label{app:thr}
``Approached'' in the main text uses 10~cm between the gripper and the bowl's body origin in the simulator at the
closest point of the episode; when both bowls come within the threshold, the nearer one counts. Of t9's 81 failed synonym episodes, 54 come within 10~cm of the target bowl and still do not complete the task (\S\ref{sec:reach}). Table~\ref{tab:thresholds} gives the counts at 6, 10 and 15~cm for the conditions
of Table~\ref{tab:reach} and the same conditions on t1 and t4. Criterion (b) of \S\ref{sec:intro} is stated at 10~cm, and the nonce-word
and landmark-phrase-removed episodes approach the bowl on the cabinet (the other bowl on t7, the target on
t9) in at most 9/150 and 5/50 episodes at 10~cm and in 68--73/150 and 33--35/50 at 15~cm.
\begin{table}[h]
  \centering\footnotesize
  \caption{Episodes approaching the target bowl / the other bowl at three thresholds (released policy).}
  \label{tab:thresholds}
  \begin{tabular}{@{}llrrrr@{}}
\toprule
Task & Instruction & $n$ & 6 cm & 10 cm & 15 cm \\
\midrule
t1 & original & 50 & 50/0 & 50/0 & 50/0 \\
t1 & synonyms (5) & 250 & 37/0 & 225/0 & 247/0 \\
t1 & landmark phrase removed & 51 & 3/0 & 36/1 & 47/1 \\
t1 & nonce words (3) & 151 & 119/0 & 149/0 & 151/0 \\
t4 & original & 50 & 34/0 & 50/0 & 50/0 \\
t4 & synonyms (5) & 250 & 167/0 & 250/0 & 250/0 \\
t4 & landmark phrase removed & 50 & 6/0 & 50/0 & 50/0 \\
t4 & nonce words (3) & 150 & 102/0 & 150/0 & 150/0 \\
t7 & original & 50 & 50/0 & 50/0 & 50/0 \\
t7 & synonyms (5) & 250 & 223/2 & 237/3 & 237/13 \\
t7 & landmark phrase removed & 50 & 0/1 & 0/4 & 0/35 \\
t7 & other landmark's word & 50 & 0/49 & 0/50 & 0/50 \\
t7 & nonce words (3) & 150 & 0/3 & 0/9 & 1/73 \\
t7 & other landmark's synonym & 50 & 0/29 & 0/41 & 0/50 \\
t9 & original & 50 & 50/0 & 50/0 & 50/0 \\
t9 & synonyms (5) & 250 & 156/0 & 223/0 & 250/0 \\
t9 & landmark phrase removed & 50 & 2/0 & 5/0 & 33/0 \\
t9 & other landmark's word & 50 & 0/50 & 0/50 & 0/50 \\
t9 & nonce words (3) & 150 & 2/0 & 4/0 & 68/0 \\
t9 & other landmark's synonym & 50 & 0/50 & 0/50 & 0/50 \\
\bottomrule
\end{tabular}

\end{table}

\subsection{Parent-VLM control}\label{app:pvlm}
The control of \S\ref{sec:reach} runs the parent model, Cosmos-Reason2-2B, with all 28 layers and
its language-model head, in fp32, on the starting frames of demonstrations 0--9 of each of t1, t4, t7
and t9 with the edited instruction, and asks which bowl the instruction names. The candidates are the phrases that locate the scene's two bowls (on t4 they share the noun and differ
only in the relation phrase, so t4's 50 cases do not test the synonym); each is scored by its summed log-likelihood under teacher forcing in both
presentation orders, and a case counts as correct when the correct bowl's margin, averaged over the
two orders, is positive. In 36 of the 200 cases the correct bowl wins in one
presentation order only. Summing favors the shorter phrase, which on the mirror pair is the correct
one on t7 and the wrong one on t9. The 200 cases are four tasks, five synonyms and ten frames; frames
of one word are not independent, and t4's five synonyms are t9's, so the cases rest on 15 distinct words. Layers 17--28 are not in the
policy.

\subsection{Finetuning recipe}\label{app:recipe}
Every finetuned policy of \S\ref{sec:repair} starts from the released checkpoint and trains for 1{,}500 steps at
learning rate $10^{-5}$, effective batch 64 (batch 2, 32 accumulation steps), cosine schedule
with 5\% warm-up, weight decay $10^{-5}$, state dropout 0.2, color jitter, with the VLM frozen and
all 537 action-head tensors (\S\ref{sec:path}) trainable, on the LIBERO-Spatial demonstrations.
The landmark-diversified and verb-diversified policies differ from the undiversified policy only
in the instruction sampled per frame; on the mirror pair the added landmark word was chosen, from a pilot run, as the synonym the released policy completed least (Table~\ref{tab:subst_cells}'s unmodified column shows the same ranking). Training seeds 101, 202 and 303 seed all of training (data order and every other random draw); nothing is
re-initialized. Checkpoints are compared tensor by tensor and the VLM (the code's \emph{backbone}) is
identical across all of them. Connector-only finetuning uses the same recipe with every action-head tensor outside the connector
frozen too; frozen modules run in evaluation mode, so the diffusion
transformer's dropout is off during that training. Each connector-only checkpoint differs from the
released one in exactly the 66 connector tensors.

\subsection{Checkpoints and versions}\label{app:ckpt}
The released checkpoint is \texttt{nvidia/GR00T-N1.7-LIBERO} (\texttt{libero\_spatial}, revision
\texttt{2ea293aa}); the parent VLM is \texttt{nvidia/Cosmos-Reason2-2B} (revision \texttt{9ce19a19}).
Compared tensor by tensor with the parent, the released policy's VLM differs in all 176 tensors of its sixteen kept
language-model layers and in 477 of its 494 backbone tensors in all; it is not a plain truncation of the parent.

\subsection{Receipts}\label{app:receipts}
Each paste or masking condition writes, on first use, the positions replaced or masked,
their tokens, (for pastes) the decoded recipient and donor instructions, and (position-group pastes before the
connector) the maximum absolute difference between recipient and donor vectors over the replaced positions. The run stops if any
field disagrees with the specification. The pre-connector image-group paste has a maximum
difference of 0.0 on both tasks, which is the check that the causal VLM's image positions do not
see the instruction.

\subsection{Open-loop measure}\label{app:openloop}
The retained fraction of a synonym instruction $c$ is $1-d(c,o)/d(s,o)$, where $d(x,y)$ is the
root-mean-square difference between the action chunks predicted for the same frame under instructions
$x$ and $y$, $o$ is the original instruction and $s$ the one with the other landmark's word; the robot state
is zeroed (its raw values before normalization) and the diffusion noise fixed. It is computed on the first frame of ten demonstrations per task, averaged
over the five synonyms, and a transplant's value is its difference from the undiversified
policy, paired by frame. The paired $t$ of \S\ref{sec:position} (nine degrees of freedom) is the mean of the ten per-frame
differences divided by their sample standard deviation over $\sqrt{10}$; the values in
\S\ref{sec:position} ($|t|\le1.6$) correspond to two-sided $p>0.1$. On the mirror pair pooled, the all-but-connector
transplant changes the score by +0.008 to +0.013 and the connector transplant by +0.204 to +0.210. On t9's four unpresented synonyms
alone, the all-but-connector transplant changes it by $-0.008$, $-0.013$ and $-0.005$ ($|t|\le1.8$).

\subsection{Cells behind the rescue and the mask}\label{app:bcells}
Table~\ref{tab:subst_cells} gives the rescue of Table~\ref{tab:locus}B word by word, with the two
position controls on the cells that have them; Table~\ref{tab:ko_cells} shows the mask of
Table~\ref{tab:locus}C with its matched-position control. For each of the two, the decision cells and the
unmodified baseline come from the same run; the rescue and the mask come from separate runs. Against
the unmodified cells of the other two runs (\S\ref{sec:stats}), the pre-connector rescue is +38.0 and
+36.0 points and the mask $-$24.0 and $-$20.4. Over all ten synonym cells (five already at 45/50 or more),
the pre-connector rescue gains +18.6 points ($z=9.22$). Cell by cell (Table~\ref{tab:ko_cells}), the mask's cost
separates on hotplate and bureau and is unresolved on the other three; chest of drawers and sideboard move by at most
two episodes.

Pasting the other landmark's vector at the landmark position after the connector (\S\ref{sec:postneck}) leaves the cost
to t9 completion unresolved with environment seed 0 (39/50 against 49/51, $z=-2.71$); with seed 100 the
cost separates (31/50 against 50/50, $z=-4.84$), though only 4 of 50 approach the other bowl. Of the
post-connector pastes into original-instruction episodes (Table~\ref{tab:locus}A), the landmark and
the rest of the instruction pasted together, read by the instruction cross-attention alone, flip 41 and 47, so a paste need not
reach both cross-attention branches. The image group alone lowers completion from 49/50 in that run
to 37/50 on t7 ($z=-3.46$) and to 43/50 on t9, unresolved. The other closed-loop cells behind
Tables~\ref{tab:locus} and~\ref{tab:repair} (position groups, landmark plus one other group, environment
seed 100, task t1, and the mask on the finetuned policies) are not shown here.
\begin{table}[h]
  \centering\scriptsize\setlength{\tabcolsep}{3pt}
  \caption{Completed/$n$ with the original instruction's landmark vectors pasted into a synonym
  episode (released policy, environment seed 0, run 1 of Table~\ref{tab:locus}B). Control positions: as many other positions
  as the landmark span, and the verb's one position; before the connector the verb precedes the landmark, so the
  causal VLM gives it the same vector in both instructions and its paste changes nothing (its cells re-measure
  the unmodified condition), while each matched position receives the average of the donor's vectors at the matched positions (\S\ref{sec:interventions}).
  On t7 the landmark is one token, so the own-vector paste of \S\ref{sec:interventions} replaces that vector with itself and changes nothing either. $^{\dagger}$Decision cells.}
  \label{tab:subst_cells}
  \begin{tabular}{@{}llrrrrr@{}}
\toprule
 & & & \multicolumn{2}{c}{Landmark position} & \multicolumn{2}{c}{Control positions (before / after)} \\
Task & Instruction & Unmodified & before connector & after connector & matched other & verb \\
\midrule
t7 & ``cooktop'' & 49/50 & 49/50 & 49/50 & -- & -- \\
t7 & ``range'' & 47/50 & 49/50 & 50/50 & -- & -- \\
t7 & ``cooker'' & 47/50 & 48/50 & 50/50 & -- & -- \\
t7 & ``hob'' & 50/50 & 49/50 & 49/50 & -- & -- \\
t7 & ``hotplate''$^{\dagger}$ & 31/50 & 47/50 & 50/50 & 31/50 / 35/50 & 28/50 / 29/50 \\
t7 & original (own vector pasted) & 48/50 & 50/50 & 49/50 & -- & -- \\
\midrule
t9 & ``cupboard'' & 45/50 & 50/50 & 45/50 & -- & -- \\
t9 & ``dresser''$^{\dagger}$ & 39/50 & 50/50 & 41/50 & -- & -- \\
t9 & ``chest of drawers''$^{\dagger}$ & 25/50 & 50/50 & 44/50 & 23/50 / 29/50 & 30/50 / 23/50 \\
t9 & ``sideboard''$^{\dagger}$ & 28/50 & 50/50 & 41/50 & -- & -- \\
t9 & ``bureau''$^{\dagger}$ & 35/50 & 47/50 & 41/50 & -- & -- \\
t9 & original (own vector pasted) & 49/51 & 49/50 & 50/50 & -- & -- \\
\bottomrule
\end{tabular}

\end{table}
\begin{table}[h]
  \centering\scriptsize
  \caption{The post-connector mask on the released policy (environment seed 0; the run of Table~\ref{tab:locus}C, separate from
  Table~\ref{tab:subst_cells}'s): completed/$n$, with
  approach in parentheses as target / other / neither.}
  \label{tab:ko_cells}
  \begin{tabular}{@{}llrrr@{}}
\toprule
Task & Instruction & Unmasked & Landmark masked & Matched other positions masked \\
\midrule
t7 & ``hotplate'' & 32/50 (37/4/9) & 13/50 (15/13/22) & 31/50 (38/3/9) \\
t7 & original & 49/50 (50/0/0) & 49/50 (50/0/0) & -- \\
t9 & ``dresser'' & 39/50 (48/0/2) & 26/50 (40/0/10) & -- \\
t9 & ``chest of drawers'' & 23/50 (40/0/10) & 21/50 (38/0/12) & 26/50 (43/0/7) \\
t9 & ``sideboard'' & 22/50 (41/0/9) & 20/50 (32/0/18) & -- \\
t9 & ``bureau'' & 38/50 (46/0/4) & 18/50 (37/0/13) & -- \\
t9 & original & 49/50 (50/0/0) & 47/50 (48/0/2) & -- \\
\bottomrule
\end{tabular}

\end{table}

\subsection{Stratified, combined and family-wise checks}\label{app:stats}
For the 99 closed-loop comparisons behind the
verdicts of the body and Appendix~\ref{app:bcells}, we computed a 95\% interval of the pooled difference (Newcombe's hybrid score
interval) and Fisher's exact $p$. Stratifying the 72 that pool cells by cell (Mantel--Haenszel difference,
Cochran--Mantel--Haenszel $z$) changes no verdict; for t9's five synonyms against the training word,
which share no strata, inflating the synonyms' variance by their between-cell dispersion gives
$z=-3.55$ instead of $-4.71$. With no true difference, $|z|\ge3$ would separate 0.27 of the 99
comparisons on average. Three separations do not survive Holm's correction at a family-wise 0.05:
the instruction cross-attention blocks against all 471 tensors in seed 101 ($z=-3.14$; Appendix~\ref{app:subsets}), the connector transplant's nonce-word completion on t9 against the diversified policy's in seed 303 ($-3.17$; Appendix~\ref{app:drift}) and the
diversified policy's masking cost in seed 303 ($-3.20$; \S\ref{sec:after}). Four more survive Holm's
but not Bonferroni's (threshold $|z|\ge3.48$): the image group's completion on t7 (Appendix~\ref{app:bcells}),
the image cross-attention blocks in seed 101 ($-3.34$) and the masking cost on the three unpresented cells in seeds
101 and 303 (\S\ref{sec:after}). Combining the three training seeds' $z$ (Stouffer) separates every
family the body reports as separated in some seeds, and two that are unresolved in every seed: the landmark-diversified connector-only policy's nonce-word completion on t9 against the diversified policy's ($-3.22$) and
sideboard's masking cost after the repair ($-4.75$). The decision-cell differences between
connector-only and full finetuning ($-0.4$, 0.0 and $-0.8$ points with the diversified instructions, 95\% intervals within $-3.5$ to $+2.4$;
$-8.8$, $-0.2$ and $+2.0$ with the original ones, seed 101's interval $-17.3$ to $-0.1$) stay unresolved combined ($-0.64$, $-0.90$).

\section{Supporting observations}\label{app:obs}
\subsection{Subsets of the other 471 tensors}\label{app:subsets}
The four subsets of \S\ref{sec:transplant} (Table~\ref{tab:subsets}) are the eight instruction
cross-attention blocks (112 tensors), the eight image cross-attention blocks (112), the sixteen
self-attention blocks (224) and the 23 input and output tensors (state and action encoders, action
decoder, action position embedding, timestep encoder and output projections). The three kinds of
blocks lie above the undiversified policy in every seed, the input and output tensors not in
every seed, but none separates from that policy ($|z|<2.9$), whereas the 471 together do ($z=5.3$ to 6.5); each subset is below the 471 in every seed ($z=-3.1$ to
$-6.8$; one fails Holm's correction, Appendix~\ref{app:stats}). We do not call the parts additive.
\begin{table}[h]
  \centering\scriptsize
  \caption{Completion after transplanting subsets of the other 471 tensors into the undiversified policy
  (same training seed), mirror-pair synonyms, in percent; training seeds 101, 202, 303 in that order.
  Unpresented: the eight synonym cells not in the finetuning instructions; presented: the two that are;
  original: the original instruction. Difference and $z$ are against the same seed's undiversified policy,
  pooled over the eight unpresented cells.}
  \label{tab:subsets}
  \setlength{\tabcolsep}{6pt}\begin{tabular}{@{}lrrrr@{}}
\toprule
Policy & Unpresented (\%) & Difference, pt ($z$) & Presented (\%) & Original (\%) \\
\midrule
undiversified & 80.8, 79.1, 81.8 & -- & 64.0, 50.0, 47.0 & 100.0, 98.0, 98.0 \\
instruction cross-attention (112) & 87.2, 85.2, 86.0 & +6.5 (2.5), +6.2 (2.3), +4.2 (1.6) & 68.0, 60.0, 62.0 & 99.0, 99.0, 98.0 \\
image cross-attention (112) & 86.8, 86.5, 83.0 & +6.0 (2.3), +7.5 (2.8), +1.3 (0.5) & 63.0, 55.0, 58.0 & 98.0, 99.0, 99.0 \\
self-attention (224) & 85.8, 82.5, 85.0 & +5.0 (1.9), +3.5 (1.3), +3.3 (1.3) & 67.0, 55.0, 56.0 & 98.0, 100.0, 98.0 \\
input and output tensors (23) & 84.0, 78.1, 81.0 & +3.3 (1.2), $-$1.0 ($-$0.3), $-$0.8 ($-$0.3) & 61.0, 44.0, 58.0 & 97.1, 98.0, 99.0 \\
all-but-connector transplant & 93.8, 94.5, 94.0 & Table~\ref{tab:repair}B & 88.2, 76.0, 82.0 & 98.0, 98.0, 98.0 \\
connector transplant & 98.5, 99.0, 99.0 & Table~\ref{tab:repair}B & 98.0, 98.0, 98.0 & 99.0, 97.0, 98.0 \\
landmark-diversified & 98.8, 99.3, 99.5 & Table~\ref{tab:repair}B & 100.0, 99.0, 96.0 & 97.0, 98.0, 98.0 \\
\bottomrule
\end{tabular}

\end{table}

\subsection{The flip pattern on finetuned policies}\label{app:flipft}
On the finetuned policies the flips largely keep the released policy's pattern (Table~\ref{tab:flip_finetuned}: three of
the flips of Table~\ref{tab:locus}A, the landmark before the connector, and the landmark and all except
the landmark after it). The pre-connector landmark position moves 49 or 50 of 50 episodes in every cell and
the post-connector complement 38--50, for both policies and all seeds. This run has no unmodified cells. On the undiversified policies, with the other landmark's vector pasted at the post-connector landmark
position, 27, 37 and 39 of 50--51 complete t9, against 50/50 in every seed for the unmodified instruction in a separate
run, the evaluation of Table~\ref{tab:repair}A--B. With no same-run baseline we report the counts without a verdict, in the direction seen on the released policy. The same paste flips
2/50, 1/51 and 2/51 t9 episodes on the undiversified policies and 15, 6 and 11 of 50 on the diversified
ones (separated from the undiversified ones in one seed, unresolved in two).

\subsection{The masking cost with the decision cells fixed}\label{app:maskfixed}
With the released policy's five decision cells fixed for every policy and seed, instead of the rule
of \S\ref{sec:stats}, the mask of Table~\ref{tab:repair}C costs 24.4, 26.2 and 27.2 points on the
undiversified policies and 6.4, 6.4 and 5.6 on the landmark-diversified ones, a difference of
+18.0, +19.8 and +21.6 ($z=3.8$, 4.4 and 4.7); the fixed cells differ from those the rule of \S\ref{sec:stats} selects only in seed 101,
where the rule drops hotplate (Table~\ref{tab:repair}). On the t7 cell, where the drift points to the other bowl, the
undiversified policies lose 46, 48 and 32 points under the mask while the diversified ones change by 0, +2 and +4;
hotplate is a word in their finetuning instructions, so on this cell, as on chest of drawers, the
smaller cost may come from finetuning on the word itself. On the diversified policies, bureau, which like sideboard is not in their
finetuning instructions, completes 43/50, 45/50 and 47/50 under the mask against 51/51, 50/50 and 49/50
without it, each unresolved; sideboard completes 43/50 under the mask against 50/50 in every seed, unresolved per seed.
The two costs start from different heights: unmasked, the cells the rule selects complete at 98.8--99.0\% on the
diversified policies and at 50.6--54.4\% on the undiversified ones, so the ratios of \S\ref{sec:after} compare
drops from different baselines.

\begin{table}[h]
  \centering\scriptsize
  \caption{Flips on the finetuned policies (original instruction; seeds ordered as in Table~\ref{tab:subsets}; episodes approaching the other bowl, of the $n$ in the Completed/$n$ column; Completed: the original
  instruction's task).}
  \label{tab:flip_finetuned}
  \begin{tabular}{@{}lllrr@{}}
\toprule
Policy & Task & Site, positions & Completed/$n$ & To other bowl \\
\midrule
undiversified & t7 & post-connector, all except landmark & 2/50, 0/50, 0/50 & 46, 50, 48 \\
undiversified & t7 & post-connector, landmark & 48/51, 46/50, 49/50 & 0, 0, 0 \\
undiversified & t7 & pre-connector, landmark & 0/50, 0/50, 0/50 & 50, 50, 49 \\
undiversified & t9 & post-connector, all except landmark & 1/50, 3/50, 2/50 & 47, 46, 47 \\
undiversified & t9 & post-connector, landmark & 27/50, 37/51, 39/51 & 2, 1, 2 \\
undiversified & t9 & pre-connector, landmark & 0/50, 0/50, 0/50 & 50, 50, 50 \\
landmark-diversified & t7 & post-connector, all except landmark & 9/50, 2/50, 5/50 & 38, 46, 45 \\
landmark-diversified & t7 & post-connector, landmark & 48/50, 48/50, 48/50 & 0, 0, 0 \\
landmark-diversified & t7 & pre-connector, landmark & 0/50, 0/50, 0/50 & 50, 50, 50 \\
landmark-diversified & t9 & post-connector, all except landmark & 0/50, 0/50, 0/50 & 50, 49, 50 \\
landmark-diversified & t9 & post-connector, landmark & 28/50, 34/50, 30/50 & 15, 6, 11 \\
landmark-diversified & t9 & pre-connector, landmark & 0/50, 0/50, 0/50 & 50, 50, 50 \\
\bottomrule
\end{tabular}

\end{table}

\subsection{Where the finetuned policies go when the landmark cannot be read}\label{app:drift}
With a nonce word or the landmark phrase removed, the landmark-diversified policy drifts to the bowl
on the cabinet on both tasks (Table~\ref{tab:bias}; 16--24 of 50 per cell on t7, 11--28 on t9), while, on t7 and t9, the undiversified and verb-diversified policies approach either bowl
in at most 6 of 50 episodes per cell. Against the fully finetuned diversified policy, the
landmark-diversified connector-only policy approaches t7's other bowl with a nonce word in 12, 4 and 10
of 50 against 24, 22 and 18, separated in one seed ($z=-2.5$, $-4.1$ and $-1.8$); on t9 its nonce-word
completions (\S\ref{sec:transplant}) against the diversified policy's (\S\ref{sec:var}) are unresolved in every seed
($|z|\le2.3$), and the connector transplant's are separated in one seed ($z=-1.6$, $-2.5$ and $-3.2$; seed 303's not after Holm's
correction, Appendix~\ref{app:stats}).
\begin{table}[!h]
  \centering\scriptsize
  \caption{Approach when the landmark cannot be read (finetuned and transplanted policies; seeds ordered as in
  Table~\ref{tab:subsets}). The cabinet's bowl is the other bowl on t7 and the target on t9.}
  \label{tab:bias}
  \begin{tabular}{@{}lllrrr@{}}
\toprule
Instruction & Task & Policy & Completed/$n$ & To target & To other \\
\midrule
nonce word & t7 & undiversified & 0/50, 0/50, 0/50 & 0, 0, 0 & 4, 1, 2 \\
nonce word & t7 & verb-diversified & 0/50, 0/50, 0/50 & 0, 0, 0 & 4, 1, 0 \\
nonce word & t7 & landmark-diversified & 0/50, 0/50, 0/50 & 0, 1, 0 & 24, 22, 18 \\
nonce word & t7 & connector transplant & 0/50, 0/50, 0/50 & 0, 0, 0 & 11, 6, 9 \\
nonce word & t7 & all-but-connector transplant & 0/50, 0/50, 0/50 & 0, 0, 0 & 8, 6, 6 \\
nonce word & t9 & undiversified & 0/50, 1/50, 0/50 & 3, 1, 0 & 0, 0, 0 \\
nonce word & t9 & verb-diversified & 0/50, 0/50, 0/50 & 1, 0, 1 & 0, 0, 0 \\
nonce word & t9 & landmark-diversified & 12/50, 10/50, 15/50 & 26, 28, 24 & 1, 0, 0 \\
nonce word & t9 & connector transplant & 6/50, 2/50, 3/51 & 13, 7, 8 & 0, 0, 0 \\
nonce word & t9 & all-but-connector transplant & 2/50, 2/51, 2/50 & 9, 4, 5 & 0, 0, 0 \\
landmark phrase removed & t7 & undiversified & 0/50, 0/50, 0/50 & 0, 0, 0 & 3, 2, 1 \\
landmark phrase removed & t7 & verb-diversified & 0/50, 0/50, 0/50 & 0, 0, 0 & 2, 2, 2 \\
landmark phrase removed & t7 & landmark-diversified & 0/50, 0/50, 0/50 & 1, 0, 0 & 16, 20, 19 \\
landmark phrase removed & t7 & connector transplant & 0/50, 0/50, 0/50 & 0, 0, 0 & 9, 4, 5 \\
landmark phrase removed & t7 & all-but-connector transplant & 0/50, 0/50, 0/50 & 0, 0, 0 & 11, 5, 7 \\
landmark phrase removed & t9 & undiversified & 1/50, 0/50, 0/50 & 6, 3, 3 & 0, 0, 0 \\
landmark phrase removed & t9 & verb-diversified & 1/50, 0/50, 0/50 & 4, 1, 1 & 0, 0, 0 \\
landmark phrase removed & t9 & landmark-diversified & 10/50, 7/50, 7/50 & 23, 11, 16 & 0, 0, 0 \\
landmark phrase removed & t9 & connector transplant & 2/50, 2/51, 0/50 & 12, 4, 4 & 0, 0, 0 \\
landmark phrase removed & t9 & all-but-connector transplant & 5/52, 3/51, 4/52 & 13, 6, 10 & 0, 0, 0 \\
landmark phrase removed & t1 & undiversified & 7/50, 7/50, 5/50 & 36, 41, 39 & 0, 0, 0 \\
landmark phrase removed & t1 & verb-diversified & 6/51, 4/51, 5/50 & 39, 35, 39 & 0, 0, 0 \\
landmark phrase removed & t1 & landmark-diversified & 33/50, 36/50, 32/50 & 50, 50, 50 & 0, 0, 0 \\
\bottomrule
\end{tabular}

\end{table}

\subsection{Other checkpoints, and destination phrases that covary with the action}\label{app:otherckpt}
The runs in this section concern phrases other than the landmark and, in the last, a change to the
demonstrations; they are reported for completeness.
On the released LIBERO-Goal checkpoint the destination phrase is followed to fixtures other than
the scene's training one (the bowl is placed at the named fixture in 297/300 episodes off the
diagonal of the task-by-destination matrix; the training task completes in 146/150 on it and 0/300
off it). Destination synonyms are followed (248/250 completed, against 54/150 with the destination phrase removed; that
control is high only on the Goal task whose destination is the cabinet, 45/50, where the bowl mostly goes without the phrase). On the Spatial checkpoint itself the destination phrase (``on the plate'') is not followed: with another
destination named, the bowl still goes on the plate (196/200; none at the named one). On the released Object checkpoint, naming another object of the scene
lowers completion of the original task to 1/500; removing or replacing the destination phrase leaves
completion high (490/501, 499/502), and an object synonym that shares no token with the trained noun
fails (1/50; synonyms that share one, 432/450). Where the policy goes was not recorded in any of these
Object runs; \citet{encodednotcontrol}, with a checkpoint they describe as finetuned on the full LIBERO
benchmark and with LIBERO's fixed initial states (their App.~A.2--A.3), find the original object still
completed in 90.9\% of such episodes. Finetuning the Spatial checkpoint on Goal demonstrations with the destination
phrase covarying with the action in $P$\% of the three bowl tasks' demonstrations (the other seven tasks keep their instructions) gives destination
following (the fraction of episodes that place the bowl at the named destination) of 0.16--0.19, 0.72--0.76, 0.84--0.90 and 0.90--0.95 at $P=0$, 25, 50 and 100 over the
three seeds. A frequency-matched shuffle of the phrases gives 0.32, against 1/3 for three destinations.

\section{Reproduction}\label{app:repro}
Measurements ran on one NVIDIA GB10 (aarch64, CUDA 13.0) with torch 2.10.0, transformers 4.57.3 and mujoco 3.3.1; one evaluation cell takes about two minutes and one finetune of 1{,}500 steps about four hours.

\end{document}